\pdfoutput=1
\documentclass[11pt]{article}
\usepackage[a4paper, margin=1in]{geometry}  % Ensures 1-inch margins

\usepackage[preprint]{acl}
\usepackage{enumitem} % Add this line in your preamble
\usepackage[compact]{titlesec} % Add this to your preamble

\titlespacing*{\section}{0pt}{1ex}{1ex} % Adjust section spacing

\usepackage{times}
\usepackage{latexsym}
\usepackage{framed} % For creating text boxes
\usepackage[T1]{fontenc}
\usepackage{booktabs}
\usepackage{tabularx} % For adjusting table width
\usepackage{fdsymbol}
\usepackage{pgfplots}
\pgfplotsset{compat=1.18}
\usepackage{pgfplotstable}
\usepackage{amsmath}
\usepackage{comment}
\usepackage{float}
\usepackage{colortbl}
\usepackage{xcolor}  % For defining colors
\usepackage{tabularx} % For clean table formatting

\usepackage[utf8]{inputenc}

\usepackage{microtype}

\usepackage{inconsolata}

\usepackage{graphicx}

\title{Afrispeech-Dialog: A Benchmark Dataset for Spontaneous English Conversations in Healthcare and Beyond}

\author{
  \textbf{Mardhiyah Sanni\textsuperscript{1,2}},  
  \textbf{Tassallah Abdullahi\textsuperscript{2,3}},  
  \textbf{Devendra D. Kayande\textsuperscript{1,2,4}},  
\\  
  \textbf{Emmanuel Ayodele\textsuperscript{1}},  
  \textbf{Naome A. Etori\textsuperscript{2,5}},  
  \textbf{Michael S. Mollel\textsuperscript{2,6}},
  \textbf{Moshood Yekini\textsuperscript{2}},
\\  
  \textbf{Chibuzor Okocha\textsuperscript{2,7}},  
  \textbf{Lukman E. Ismaila\textsuperscript{2,8}},
   \textbf{Folafunmi Omofoye\textsuperscript{2,9}},
\\  
 \textbf{Boluwatife A. Adewale\textsuperscript{2}},  
  \textbf{Tobi Olatunji\textsuperscript{1,2,10$\spadesuit$}}  
\\
 \textsuperscript{1}Intron,  
  \textsuperscript{2}BioRAMP,  
  \textsuperscript{3}Brown University  
  \\  
  \textsuperscript{4}Indian Institute of Information Technology Allahabad,  
  \textsuperscript{5}University of Minnesota-Twin Cities  
  \\  
  \textsuperscript{6}University of Glasgow,  
  \textsuperscript{7}University of Florida,  
  \textsuperscript{8}Johns Hopkins University  
  \\  
  \textsuperscript{9}University of North Carolina at Chapel Hill,  
  \textsuperscript{10}Georgia Institute of Technology  
  \\  
  \texttt{tobi@intron.io}  
}

\begin{document}
\maketitle
\footnotetext[1]{Authors marked with $\spadesuit$ are senior authors.}
\begin{abstract}
Speech technologies are transforming interactions across various sectors, from healthcare to call centers and robots, yet their performance on African-accented conversations remains underexplored. We introduce Afrispeech-Dialog, a benchmark dataset of 50 simulated medical and non-medical African-accented English conversations, designed to evaluate automatic speech recognition (ASR) and related technologies. We assess state-of-the-art (SOTA) speaker diarization and ASR systems on long-form, accented speech, comparing their performance with native accents and discover a 10\%+ performance degradation. Additionally, we explore medical conversation summarization capabilities of large language models (LLMs) to demonstrate the impact of ASR errors on downstream medical summaries, providing insights into the challenges and opportunities for speech technologies in the Global South. Our work highlights the need for more inclusive datasets to advance conversational AI in low-resource settings.
%\textcolor{red}{
%This document is a supplement to the general instructions for *ACL authors. It contains instructions for using the \LaTeX{} style files for ACL conferences. The document itself conforms to its own specifications and is an example of what your manuscript should look like.These instructions should be used for papers submitted for review and for final versions of accepted papers.}
\end{abstract}

\section{Introduction}

Deep Learning approaches have revolutionized and yielded significant performance gains across several natural language tasks \citep{bharadiya2023comprehensive}, especially for high-resource languages \cite{radford2023robust}. While conversational speech recognition has continued to make significant strides in task automation in different domains such as medical \citep{biswas2024intelligent, abacha2023empirical, yim2023aci}, voice assistants \citep{pasandi2022evaluation, mani2020towards}, call center \citep{plaza2021call}, and robotics \citep{skantze2021turn}, much of the research in Automatic Speech Recognition (ASR) has focused on monolingual speech with native accents \cite{aksenova2022accented}, with considerable performance gaps in diverse linguistic, low resource, and accented contexts \cite{radford2023robust, olatunji2023afrinames}. %This gap is complicated by limited annotated open-access datasets featuring African-accented conversations and limited computational resources. 

%Code-switching is a phenomenon in African linguistic environments where speakers alternate between languages during conversations. This presents a unique challenge that further complicates the development and evaluation of ASR systems that transcribe and process long-chain spontaneous dialogue in African-accented languages accurately. And it is perhaps no wonder that studies have shown the staggering generalization performance of state-of-the-art (SOTA) models \cite{hinsvark2021accented} in both general \cite{adelani2022masakhaner, olatunji2023afrinames, ogun20241000, tonja2024ethiollm} and clinical closed \cite{afonja2024performant, olatunji2023afrispeech} domain tasks on African-accented languages. 

In Anglophone African contexts, regional accents further complicate the development of ASR systems resulting in poor generalization from state-of-the-art (SOTA) models \cite{hinsvark2021accented} in both general \cite{adelani2022masakhaner, olatunji2023afrinames, ogun20241000, tonja2024ethiollm} and medical  \cite{afonja2024performant, olatunji2023afrispeech} contexts. In more developed countries, Performant Speech recognition models are particularly useful in overburdened healthcare systems, where they can help to reduce the documentation workload for overwhelmed clinicians \cite{afonja2024performant, olatunji2023afrispeech}. 

% In developed countries, effective speech recognition models are particularly valuable in overburdened healthcare systems, where they help reduce the documentation workload for overwhelmed clinicians

%The integration of deployable model-based ASR systems into clinical workflow is gaining traction globally including Africa \footnote{https://techcabal.com/2023/05/29/intron-health-brings-ai-to-african-healthcare/}\footnote{https://techtrendske.co.ke/2024/09/23/google-supports-jacaranda-healths-use-of-gen-ai-for-mothers-and-babies/}. Its efficiency is additionally challenged by governmental regulations around data-sharing and concerns about the monetization of sensitive data. Hence, a need for a representative, diverse, anonymized, and readily available dataset accessible to all common interests for ethical use.

Recent advancements in medical conversation transcription and summarization using LLMs have led to wide adoption of these technologies in hospitals in developed countries \citep{michalopoulos2022medicalsum, yim2023aci} helping to improve clinical documentation quality and productivity \citep{galloway2024impact}. Given the heavy patient loads \cite{baye2020nurses,etori2023we} as well as the recent integration of AI/NLP/ASR systems into clinical workflows in African clinics \footnote{https://techcabal.com/2023/05/29/intron-health-brings-ai-to-african-healthcare/}\footnote{https://techtrendske.co.ke/2024/09/23/google-supports-jacaranda-healths-use-of-gen-ai-for-mothers-and-babies/}, performant ambient transcription and summarization systems are highly desired. However, the generalizability of such systems to African contexts remains underexplored \cite{afonja2024performant}. 

%AfriSpeech-Dialog introduces a dataset consisting of approximately 50 simulated medical and non-medical conversations with African accents, emphasizing the inclusion of code-switching across multiple languages commonly spoken on the continent. The dataset aims to evaluate the performance of state-of-the-art (SOTA) speaker diarization models \cite{park2022review, zhang2019fully} and compare the capabilities of multilingual ASR systems (e.g., Whisper, Conformer, MMS, xlsr) on long-form accented speech. By assessing the performance of these models and technologies on African-accented conversations, AfriSpeech-Dialog contributes valuable insights into the robustness of ASR systems in underrepresented linguistic environments. Furthermore, this research explores multi-agent summarization of medical and non-medical conversation transcripts, laying the groundwork for more inclusive, globally adaptable ASR and NLP technologies in healthcare and beyond.

%This study succeeds our previous works \cite{afonja2024performant, ogun20241000}, which focused on unveiling and the need for improvement of SOTA ASR models on African accented languages in clinical and non-clinical settings, 

To tackle this problem, our contributions are as follows: 

\begin{enumerate}
    \item AfriSpeech-Dialog a novel 7hr dataset of 50 long-form simulated African-accented English medical and non-medical conversations 
    \item Benchmark the performance of SOTA speaker diarization, speech recognition, and LLM-based  summarization on African-accented conversations 
\end{enumerate}

We lay the groundwork for more inclusive  ASR and NLP technologies in healthcare and beyond.

\section{Related Work}

\subsection{ASR in Medical Conversations and Summarization}
The role of ASR in medical documentation has grown significantly, particularly in telehealth and in-person patient-physician consultations \cite{korfiatis2022primock57, galloway2024impact, michalopoulos2022medicalsum, yim2023aci}. Accurate ASR in medical dialogue is critical, as transcription errors can lead to incorrect medical records. Several datasets have been developed to facilitate the study of ASR in medical contexts. PriMock57 \cite{korfiatis2022primock57} provides primary care mock consultations in European English accents; 
\citet{fareez2022dataset} evaluates medical ASR on simulated patient-physician medical interviews with a focus on respiratory cases; \citet{enarvi2020generating} experiments automatically generated medical reports from diarized doctor-patient surgery conversations. %Their unsupervised approach (mimics machine translation and summarization but differs in the homogeneity of the source-target language, reasoning over long span of source sentences, and occurrence of incomplete and irrelevant information) reveals better performance and scalability of Transformers compared to RNN models in ROGUE-L and medical fact extractors F1 relative error rate across each section of the dataset. 
\citet{le2024real} proposed a real-time speech summarization system for medical conversations with the ability to generate summaries for every (local) and end of (global) utterances, eliminating the need for continuous update and revision of the summary state. 

These datasets primarily focus on non-African accents and therefore do not account for the challenges specific to African-accented medical speech.

\subsection{Non-medical conversational ASR} 
\citet{pkezik2022diabiz} released DiaBiz, an Annotated Corpus of over 400hrs of Polish call center dialogs. Other conversational, parliamentary, or oratory datasets like AMI \cite{carletta2005ami}, Earnings22 \cite{del2022earnings}, Voxpopuli \cite{wang-etal-2021-voxpopuli} have gained popularity on public ASR benchmarks. Conversational ASR has also been explored in other domains such as call centers \citep{plaza2021call}, and robotics \citep{skantze2021turn}. However, these datasets lack representation of African-accented speech.

\subsection{African Accented ASR}
There has been growing interest in developing ASR systems that cater to African languages; for example, \citet{yilmaz2018building} developed a multilingual ASR system for code-switched South African speech. 
% East African ASR \cite{elamin2023multilingual}.
Multilingual ASR systems, such as EVI dataset \cite{spithourakis2022evi}, offer a strong foundation for developing similar models in African contexts where data scarcity hinders progress. \citet{olatunji2023afrispeech} released a pan-African accented English dataset for medical and general ASR. While these datasets focus on single-speaker speech recognition, AfriSpeech-Dialog is the first African-accented English conversational dataset spanning medical and non-medical domains, enabling additional tasks like diarization and summarization.

\subsection{Speaker Diarization in Multi-Speaker Conversations}
To increase the efficiency of NLP/ASR systems, enormous contributions were made to researching the integration of speaker diarization (SD) into its pipeline. \citet{serikov2024proceedings} provides a comparative analysis of SD models - Pyannote \cite{bredin2020pyannote}, CLEVER\footnote{https://www.oxfordwaveresearch.com/products/cleaver/}, and NVIDIA NeMo \cite{harper2019nemo} on 20 different German dialects for diarization and identification (DI) task. NVIDIA NeMo performs slightly better with a competitive performance due to its multiscale segmentation for identifying and removing shorter segments. On a similar DI task, \cite{chua2023merlion} also benchmarked the performance of multilingual ASR models in open and closed tracks on the challenging MERLIon CCS English - Mandarin datasets - an extracted spontaneous and codeswitching parent-child conversation speeches. However, SD for African-accented conversations remains underexplored. Benchmarking SOTA SD models on AfriSpeech-Dialog reveals their limitations in this setting.

\section{Methodology}

\begin{figure*}
    \centering
   \includegraphics[width=1.0\linewidth]{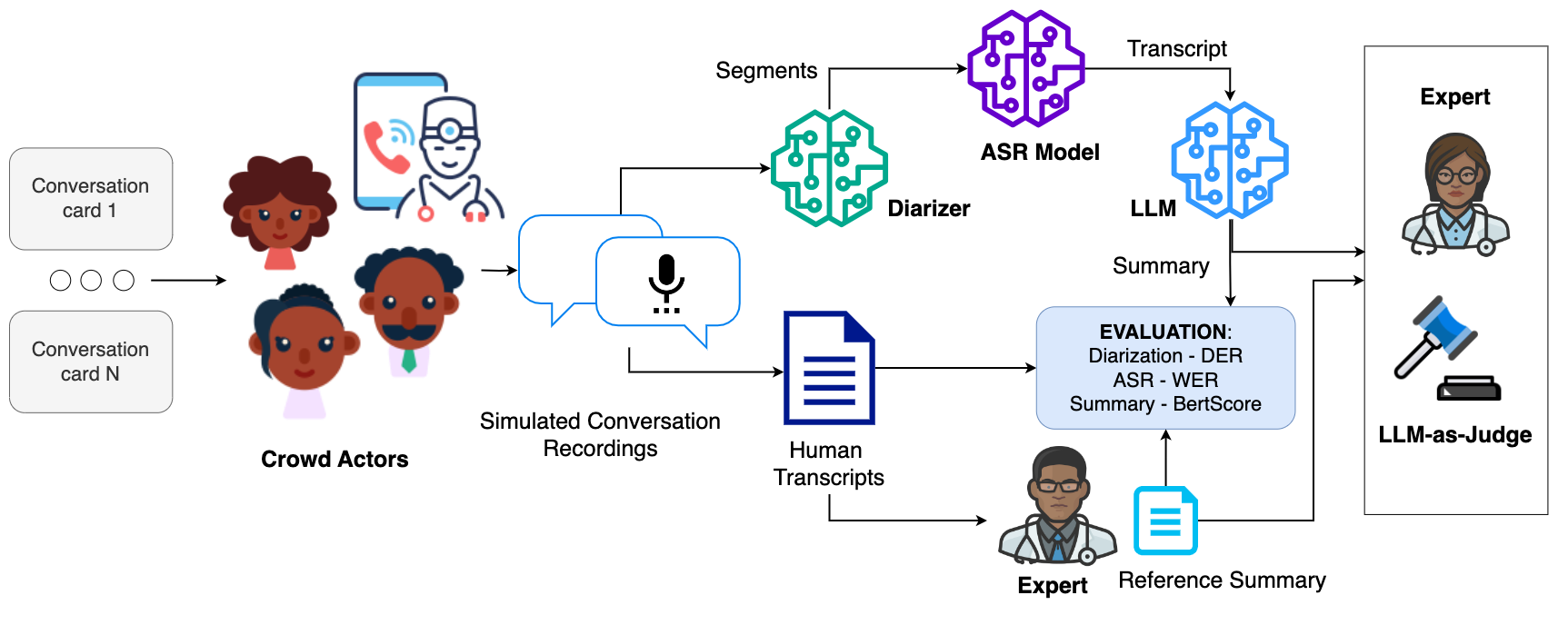}
    \caption{AfriSpeech Dialog: Dataset and Benchmarking Pipeline}
     \label{fig:methodology}
\end{figure*}

Figure \ref{fig:methodology} shows an overview of the dataset creation and benchmarking process, illustrating how AfriSpeech-Dialog supports tasks like speaker diarization, ASR, and summarization. Below, we describe the dataset creation process and the evaluation of state-of-the-art (SOTA) models on these tasks to demonstrate their applicability and highlight challenges in African-accented conversational speech.

\subsection{Dataset Statistics}
Here, we outline our dataset creation process.
\begin{table}[ht] %htb
\centering
\small
\renewcommand{\arraystretch}{1.1} % Adjust row height for better readability
\setlength{\tabcolsep}{6pt} % Adjust column spacing for better layout
\begin{tabular*}{\columnwidth}{@{\extracolsep{\fill}} l c c}
\toprule  % Bold top horizontal line
\textbf{} & \textbf{Medical} & \textbf{General} \\  
\midrule  % Single horizontal line for header separation
\textbf{Counts}           & 20              & 29                  \\  
\textbf{Timestamped Counts} & 9               & 21                  \\  
\textbf{Avg. Num. of Turns}     & 78.6            & 30.55               \\  
\textbf{Total Duration (hours)}    & 2.07            & 4.93                \\  
\textbf{Avg. Word Count}          & 725.3           & 1356.83             \\  
\textbf{Num. of Countries}          & 1           & 3             \\  
\textbf{Num. of Accents}          & 6          &      8       \\  
\textbf{Gender (M,F)}          & (14, 26)          &      (25, 33)       \\  
\bottomrule  % Bold bottom horizontal line
\end{tabular*}
\caption{\textit{Statistics of the \textbf{medical} and \textbf{non-medical} datasets.}}
\label{table:dataset_stats}
\end{table}

\subsubsection{Collecting Conversations}
%To address the need for conversational datasets that reflect African-accented English, we collected a dataset to bridge this gap. This dataset contains both medical and general domain conversations. 
We recorded simulated virtual and in-person medical and non-medical conversations from African medical and non-medical crowdworkers on the Intron Platform \footnote{https://speech.intron.health} similar to the process described in \citet{olatunji2023afrispeech}. Each conversation began with speakers providing consent, and any identifiable information in the consent segment was removed. 

% Each conversation began with both speakers giving consent to record. Because real names were sometimes spoken in the consent segment, the first 10-15 seconds were removed to protect the identities of participants.

For medical conversations, following the process described in \citet{fareez2022dataset} and \citet{korfiatis2022primock57}, clinical experts prepared "patient cards" with African-context disease conditions and demographics. Doctor and Patient actors included medical professionals (e.g. doctors, nurses) familiar with Objective Structured Clinical Examinations (OSCE), a widely used assessment in medical education that simulates doctor-patient interactions \cite{fareez2022dataset}. Each patient actor was provided with a detailed "patient card" that included information on their condition, demographics, and medical history, as shown in Table \ref{tab:patient_card}. Consistent with OSCE format, patient cards were hidden from doctor actors to facilitate a more natural consultation.

\begin{table*}[t]
\centering
\small
\renewcommand{\arraystretch}{1.2} % Adjust row height for better readability
\setlength{\tabcolsep}{10pt} % Adjust column spacing
\begin{tabularx}{\textwidth}{l X}
\toprule  
\rowcolor{gray!20} % Light gray header row
\textbf{Condition} & \textbf{Malaria} \\  
\midrule  
\textbf{Demographic (Age, Gender)} & 32-year-old Female  \\  
\rowcolor{gray!10} % Light blue background for emphasis
\textbf{Presenting Complaint} & Fever and chills (2 days) \\  
\textbf{History of Presenting Complaint} &  
\begin{itemize}
    \item \cellcolor{yellow!15} Fever for 2 days (High grade, not relieved by medication)  
    \item Chills for 2 days (Intermittent, severe)  
    \item Headache for 2 days (Generalized, throbbing, 7/10 in severity)  
    \item Fatigue and general body weakness  
    \item No cough, diarrhea, vomiting, or urinary symptoms  
    \item Patient lives in malaria-endemic area; no recent travel history  
\end{itemize} \\  
\rowcolor{gray!10} % Light gray row for subtle contrast
\textbf{Past Medical History (PMH)} & No chronic disease or surgery \\  
\textbf{Family History} & No family history of similar illness \\  
\textbf{Social History} & Does not drink alcohol or smoke \\  
\rowcolor{gray!10} % Light blue for alternating row
\textbf{Allergy History} & No known allergies \\  
\bottomrule  
\end{tabularx}
\caption{\textit{Example \textbf{Patient Card} for Medical Conversations.}}
\label{tab:patient_card}
\end{table*}

For general domain conversations, participants engaged in open discussions based on "topic cards" prepared by a team of reviewers. Each card contained a conversation topic, a brief description, and two discussion prompts to guide the conversation. The pair of participants (actors) had prior access to the cards and were advised to read through and understand them before starting the conversation. Table \ref{tab:conversation_card} shows a sample topic card.

The conversation recordings were stored as mono-channel, 16-bit wav files, with a 48 kHz sampling rate. A team of clinician reviewers reviewed the conversations and selected a high quality subset for this release. The dataset was collected across three African countries—Nigeria, Kenya, and South Africa. The speakers represent a diverse range of accents (11 in total): Hausa, Isoko, Idoma, Urhobo, Ijaw, Yoruba, Swahili, Sesotho, Igbo, Igala, and Ebira.

\begin{table*}[t]
\centering
\small
\renewcommand{\arraystretch}{1.2} % Adjust row height for readability
\setlength{\tabcolsep}{10pt} % Adjust column spacing
\begin{tabularx}{\textwidth}{X}
\toprule  
\rowcolor{gray!20} % Light gray background for header
\textbf{Topic: Cyberbullying} \\  
\midrule  
\rowcolor{green!10} % Light blue background for emphasis
\textbf{Overview:} Cyberbullying is a form of bullying that takes place online or through electronic communication.  
It involves using technology (social media, text messages, online forums) to intimidate or humiliate someone.  
Examples include:
\begin{itemize}
    \item Spreading rumors
    \item Sending hurtful messages
    \item Sharing embarrassing information without consent
\end{itemize} \\  
\rowcolor{gray!10} % Light gray row for subtle contrast
\textbf{Discussion Prompts:}  
\begin{itemize}
    \item What steps do you take to protect yourself from cyberbullying?
    \item Do you think social media has effective policies in place, or could they improve?
\end{itemize} \\  
\bottomrule  
\end{tabularx}
\caption{\textit{Sample \textbf{Conversation Card} for General Conversations.}}
\label{tab:conversation_card}
\end{table*}

%%%%%%%%%%%%%%%%%%%%%

% \begin{table*}[t]
% \centering
% \small
% \resizebox{\textwidth}{!}{%
% \begin{tabular}{|l|}
% \hline
% \textbf{Topic: Cyberbullying} \\ \hline
% \textbf{Overview:} Cyberbullying is a form of bullying that takes place online or through electronic \\ 
% communication. It involves using technology (social media, text messages, online forums) to \\ 
% intimidate or humiliate someone. Examples include spreading rumors, sending hurtful messages,\\ 
% and sharing embarrassing information without consent. \\ \hline
% \textbf{Discussion Prompts:} \\ 
% 1. What steps do you take to protect yourself from cyberbullying? \\ 
% 2. Do you think social media has effective policies in place, or could they improve? \\ \hline
% \end{tabular}
% }
% \caption{Sample Conversation Card for General Conversations}
% \label{tab:conversation_card}
% \end{table*}

% All recorded medical or general conversations lasted between 2 to 15 minutes. The dataset contains 49 conversations (20 medical and 29 general), with a total duration of approximately 7 hours. There was no unified setup for recordings; participants used mobile devices to record their conversations and uploaded them to our speech collection platform. Given that many recordings were done in environments where participants were in the same room using a single mobile phone, audio was captured on the same track for all speakers. This may pose challenges for audio segmentation but preserves the natural dynamics of real-world conversations.

\subsubsection{Recording Characteristics}
Dataset statistics are summarized in Table \ref{table:dataset_stats}.
The dataset features two speakers in each conversation in both the medical and general domains. Medical conversations were more structured, with doctors asking direct questions and patients responding, resulting in a more formal exchange. General conversations were more relaxed, with spontaneous discussions on various topics.
Overlapping speech occurs occasionally but is usually brief, involving short interjections like "yes" or "okay." Disfluencies and some code-switching reflect the natural flow of African English speakers. General conversations have fewer speaker turns but a higher average word count compared to medical ones, as speakers tend to talk longer to express their thoughts. These characteristics make the dataset valuable for testing speaker diarization and transcription models on African-accented speech.

%%%Original Table 3: statistics of the datasets%%%%%
% \begin{table}[ht]
% \centering
% \small
% \begin{tabular}{|l|c|c|}
% \hline
% \textbf{} & \textbf{Medical} & \textbf{General} \\ \hline
% \textbf{Counts}           & 20              & 29                  \\ \hline
% \textbf{Timestamped Counts} & 9               & 21                  \\ \hline
% \textbf{Avg. Num. of Turns}     & 78.6            & 30.55               \\ \hline
% \textbf{Total Duration (hours)}    & 2.07            & 4.93                \\ \hline
% \textbf{Avg. Word Count}          & 725.3           & 1356.83             \\ \hline
% \textbf{Num. of Countries}          & 1           & 3             \\ \hline
% \textbf{Num. of Accents}          & 6          &      8       \\ \hline
% \textbf{Genders (M,F)}          & (14, 26)          &      (25, 33)       \\ \hline
% \end{tabular}
% \caption{Statistics of the medical and non-medical datasets.}
% \label{table:dataset_stats}
% \end{table}

\subsubsection{Transcription Process} 
All conversations were manually transcribed by five professional annotators selected from top-performing contributors on the Intron Platform. Annotators were instructed to annotate speaker turns and insert timestamps for each turn, with all annotators required to be familiar with medical terminology. To ensure quality, clinician reviewers evaluated a random 20\% of each annotator’s work for accuracy, with at least 90\% correctness threshold required for inclusion. Contributors and annotators were paid \$10–\$20 per hour depending on task complexity and clinical experience. The dataset is released under a CC-BY-NC-SA 4.0 License.

\subsection{Speaker Diarization}
To benchmark diarization performance on this dataset, we selected three recent high-performing neural diarization models:
\begin{itemize}
    \item \textbf{Pyannote} \cite{pyannote}: This model leverages a pre-trained neural network that computes x-vector embeddings for speaker diarization. It uses an ECAPA-TDNN \cite{desplanques2020ecapa} architecture for speaker embeddings, shown to improve speaker separation in diarization tasks.
    
    \item \textbf{Reverb diarization v2} \cite{reverb_v2}: This model is an extension of Pyannote 3.0, fine-tuned on 26,000 hours of annotated speech. However, this model uses WavLM \cite{wavlm} instead of the SincNet \cite{sincnet} features in the Pyannote 3.0 basic model.  %reported that this model outperforms Pyannote's word diarization error rate by 22.5\%, making it a promising candidate for speaker diarization in complex conversational data.
    \item \textbf{Titanet}: This Nvidia diarization pipeline uses MarbleNet \cite{marblenet} for voice activity detection and Titanet-L \cite{titanet} for embedding extraction. Titanet uses a 1D depth-wise separable convolutions with Squeeze-and-Excitation (SE) layers with global context, followed by a channel attention-based statistics pooling layer to map variable-length utterances to fixed-length embeddings (t-vectors) \cite{titanet}. %The pipeline is configured using a clustering-based diarization approach with parameters tailored to estimate the number of speakers and improve clustering accuracy.
\end{itemize}

% In addition to these neural models, we also benchmarked popular commercially available speaker diarization models - AssemblyAI, Deepgram's Nova 2, Soniox’s en\_v2 model. These models were chosen due to their use in prior research \cite{sound_of_healthcare,} to handle conversational datasets.

%\subsubsection{Preprocessing}
%Prior to applying diarization, the audio data underwent a few preprocessing steps. A portion of the dataset (30 samples) contains timestamps marking the start and end of each speaker’s utterance, along with speaker tags. For the 30 timestamped audios, we extracted the timestamps and speaker tags and converted them into Pyannote’s segment format to create ground-truth reference segments for evaluation. No additional silence removal or segmentation of long audio files was performed.

\subsubsection{Evaluation Metrics}
We evaluated diarization performance using the \textbf{Diarization Error Rate (DER)} \cite{doddington2000nist}. DER quantifies the percentage of time that speakers are misattributed or missed in the diarization output. % and is computed as:

%\[
%DER = \frac{FA + MISS + ERROR}{TOTAL TIME}
%\]

%where:
%\begin{itemize}
%    \item \textbf{FA (False Alarm)} is the time when a speaker is falsely detected.
%    \item \textbf{MISS} represents missed speech segments.
%    \item \textbf{ERROR} accounts for incorrect speaker assignments, also referred to as speaker confusion.
%    \item \textbf{TOTAL TIME} is the total duration of the audio.
% \end{itemize}

We used the \textit{pyannote.metrics} library \footnote{https://pyannote.github.io/pyannote-metrics/} to calculate DER for each recording and computed the absolute DER for the entire dataset. The optimal mapping between the reference and hypothesis labels was obtained using the Hungarian algorithm \cite{kuhn1955hungarian}, ensuring an accurate alignment. We report DER on medical and general domain conversations.

\subsection{Automatic Speech Recognition (ASR)}
% (I could not fined parakeet paper hence cited NeMo toolkit)
To benchmark ASR performance, we compare SOTA open source pretrained models, Whisper \cite{radford2023robust}, Distil-Whisper \cite{gandhi2023distil}, Nvidia Parakeet \cite{harper2019nemo}, Canary \cite{puvvada2024less}, MMS \cite{pratap2024scaling} and Wav2vec2 \cite{baevski2020wav2vec}.

%from the Hugging Face OpenASR Leaderboard\footnote{https://huggingface.co/spaces/hf-audio/open-asr-leaderboard}\cite{open-asr-leaderboard}. We selected the following models from the open-source domain: \textbf{facebook/wav2vec2-large-960h}\cite{baevski2020wav2vec}, \textbf{facebook/mms-1b-all}\cite{pratap2024scaling}, \textbf{nvidia/parakeet-tdt-1.1b}, \textbf{nvidia/parakeet-rnnt-1.1b}, \textbf{nvidia/parakeet-ctc-1.1b} from \cite{harper2019nemo} , \textbf{nvidia/canary-1b}\cite{puvvada2024less}, \textbf{openai/whisper-large-v3}, \textbf{openai/whisper-large-v3-turbo},\textbf{openai/whisper-large-v2}, \textbf{openai/whisper-medium} from \cite{radford2023robust}, \textbf{disitil-whisper/distil-large-v3} and \textbf{disitil-whisper/distil-large-v2} from \cite{gandhi2023distil}.
% We also chose some third-party models based on their claimed SOTA performance: \textbf{soniox en-v2}, \textbf{assemblyai}, and \textbf{deepgram Nova 2}. 
% By using a combination of open-source pre-trained models and third-party private models, we aim to capture performance differences across varying acoustic conditions and conversational contexts.
%Selecting open-source models for benchmarking on a dialog dataset offers transparency and reproducibility. These models, often optimized on diverse, real-world speech data, provide competitive baselines for dialog-specific tasks such as handling speaker variation, natural speech, and overlapping conversations. 

Open-source models offers transparency and reproducibility. They are often trained on diverse, real-world speech data, provide competitive baselines for dialog-specific tasks such as handling speaker variation, spontaneous speech, and overlapping speakers.

\subsubsection{Preprocessing}
%The original transcripts contained timestamps and speaker tags. We removed these items from the text as they are unnecessary for the ASR task. For audio processing, the files were not changed, and neither was silence removal nor speech enhancement. As the audio files were very long and exceeded the context lengths of most of the models, the audio files were chunked into segments of 30 seconds, and each of these chunks was passed for transcription, and the output text was aggregated.

The original transcripts contained timestamps and speaker tags. We removed these items from the text as they are unnecessary for the ASR task. Long form audio recordings exceeded the context length of most ASR models. They were therefore chunked into 30 second segments for inference and transcript segments returned were concatenated. %For audio processing, the files were not changed, and neither was silence removal nor speech enhancement. As the 

\subsubsection{Evaluation Metrics}
ASR performance was evaluated using \textbf{Word Error Rate (WER)}. WER measures the total number of insertions, deletions, and substitutions in the predicted text with respect to the total number of words in the reference text. %The formula for WER is given as:

\begin{comment}
\[
\text{WER} = \frac{S + D + I}{N}
\]

where:
\begin{itemize}
    \item \(S\) is the number of substitutions,
    \item \(D\) is the number of deletions,
    \item \(I\) is the number of insertions, and
    \item \(N\) is the total number of words in the reference.
\end{itemize}
\end{comment}

\subsection{Medical Conversation Summarization}
%Model choices, rationale
%Amina, Naome, Micheal
Nine LLMs (large, small, open, closed, general, and biomedical) were benchmarked for summarizing doctor-patient dialogues. Each LLMs was presented a human conversation transcript and was prompted (Appendix section \ref{sec:summ_prompt}) to generate a detailed summary.

\paragraph{Closed-source general LLMs:} OpenAI (GPT-4o, GPT-3.5-turbo) \cite{achiam2023gpt}, and Anthropic Claude-3-Sonnet\cite{anthropic2023claude} represent leading general-purpose commercial LLMs.

\paragraph{Open-source small general LLMs:}Meta-Llama-3.1-8B-Instruct, Meta-Llama-3.2-3B-Instruct \cite{dubey2024llama}, Microsoft-Phi-3-mini-4k-instruct \cite{abdin2024phi}, and Google-Gemma-2-9b-it \cite{team2024gemma} selected for their instruction-following abilities or multilingual support, which is essential for code-switching.

\paragraph{Biomedical open-source LLMs:} m42-health-Llama3-Med42-8B\cite{med42v2}, johnsnowlabs-JSL-MedLlama-3-8B-v2.0, selected for their biomedical adaptation.
% OpenMeditron-Meditron3-8B\cite{chen2023meditron}
Examples of generated summaries are provided in Appendix section \ref{sec:summ_examples}.

\subsubsection{Quantitative Evaluation}
We used the BERTScore \cite{zhang2019bertscore} to evaluate the quality of the LLM-generated summaries against the expert-generated reference summaries. Although BERTScore is widely used, studies like \cite{hanna2021fine} have shown its limitations, particularly in capturing fine-grained semantic nuances and penalizing stylistic differences. %This may lead to a lower correlation with human judgments in capturing the nuanced semantics of medical dialogues. 

\subsubsection{Qualitative Evaluation}
To address the limitations of BERTScore, we complement it with two qualitative evaluation approaches (Human and LLM-as-Judge) where summaries were evaluated on a 5-point scale adapted from \citealp{zheng2023judging, liu2023benchmarking}), 1 (worst) to 5 (best) on the following six criteria: recall of the diagnosis, accuracy of the treatment plan, avoidance of false or fabricated information, clarity and structure, and inclusion of important positive and negative clinical details. If any criterion (e.g., treatment plan) was absent from the conversation (transcript), a score of 0 was to be assigned for that criterion. Detailed evaluation criteria are in the Appendix Section \ref{sec:human_Eval}.

\paragraph{LLM-as-Judge}
Consistent with the growing trend in recent studies \cite{zheng2023judging, liu2023benchmarking} we used generative models for automated summary evaluation. We used the OpenAI's "o1" model \cite{temsah2024openai} prompting based on the criteria mentioned above. Detailed verbatim prompts and the method for computing accuracy scores are provided in Appendix Section \ref{sec:llm_eval_prompt} and \ref{sec:llm_metric}.

\paragraph{Human Evaluation}
 In a blind study, we randomly present pairs of human transcripts and LLM- or human-generated summaries to a team of 4 clinical experts. The experts compared the information available from summaries to conversation transcripts using the 6 criteria listed above \cite{kanithi2024medic, singhal2023large, wang2023automated}. Each summary was independently rated by 2 experts.

%The LLM-generated and human-derived summaries were randomly assigned to 10 medical experts for evaluation. The expert evaluators, blinded to the source of the generated summaries (human vs. LLM), rated each summary on a scale of 1 (worst) through 5 (best) on the following six criteria: recall of the diagnosis, accuracy of the treatment plan, avoidance of false or fabricated information, clarity and structure, and inclusion of both positive and negative clinical details. A score of 0 was given for any criterion not addressed in the transcript. Detailed evaluation criteria can be found in the Appendix.

\subsubsection{Error Propagation on  Cascading Models}
Since real-world conversation summarization systems rely on imperfect ASR transcripts and accented medical ASR transcription is challenging for several ASR systems \cite{afonja2024performant}, we further evaluated summaries generated based on predicted (machine) transcripts to determine if there was a drop in quality when compared with summaries generated based on human transcripts \cite{giuseppe2021lexical}. We measured summary quality using LLM-as-Judge.

\section{Experiments}
% Hyperparameters, reproducibility, type of GPUs, training time

\subsection{Diarization}
%For the diarization experiments, the open-source models were run using the publicly released checkpoints from HuggingFace \cite{huggingface}, with no modifications to their hyperparameters. All models had no prior information about the speech segments or the number of speakers; instead, these parameters were inferred during the diarization process.
We download and run inference using publicly released checkpoints from Hugging Face \cite{wolf2020transformers}, with default hyperparameters. %All models had no prior information about the speech segments or the number of speakers; instead, these parameters were inferred during the diarization process.
We set the collar to 0.0, meaning no margin was allowed around speaker transitions, ensuring that even short overlaps (e.g., "yes" or "okay") were evaluated directly without any tolerance. Overlapping speech was also not excluded from the evaluation.

We ran inference on a single Nvidia T4 GPU. Inference for Pyannote and Reverb took approximately 2 hours while the Titanet took about 30 minutes. Results represent single runs. %The experimental setup did not involve any fine-tuning, as the models were only evaluated using their pretrained versions.

%Overlapping speech was not excluded from the evaluation, and we configured all models to process the entire dataset without skipping overlaps. 

\subsection{ASR}
%For ASR benchmarking on the dataset, the open-source 
Models were downloaded from publicly available huggingface \cite{wolf2020transformers} checkpoints with default hyperparameters and the default generation configuration was used. We ran inference on Nvidia T4 GPUs. The inference required an average of around 30 minutes for the whole dataset for the open-source models. Results represent single runs.
% while around 10-15 minutes for the third-party models.
%We did not fine-tune or give any context to the models during inference.

\begin{table*}[!ht]
\centering
\small
\renewcommand{\arraystretch}{1.1} % Adjust row height for better readability
\setlength{\tabcolsep}{6pt} % Adjust column spacing for better layout
\begin{tabular*}{\textwidth}{@{\extracolsep{\fill}} l c c c c c}
\toprule  % Bold top horizontal line
\textbf{Model} & \textbf{DER (\%)} & \textbf{Med DER (\%)} & \textbf{Gen DER (\%)} & \textbf{AMI DER (\%)} & \textbf{DIHARD DER (\%)} \\  
\midrule  % Single horizontal line for header separation
Titanet-L  & \textbf{16.27} & 34.64 & \textbf{12.28} & (1.89) & - \\  
Pyannote   & 21.30 & \textbf{31.46} & 19.09 & 24.8 (4.6) & 34.4 \\  
Reverb     & 26.87 & 58.04 & 20.10 & - & - \\  
\bottomrule  % Bold bottom horizontal line
\end{tabular*}
\caption{\textit{Diarization Error Rate (DER) for all 30 audios, with detailed results for the \textbf{Medical (Med. DER)} and \textbf{General (Gen. DER)} subsets.  
The \textbf{AMI DER} and \textbf{DIHARD DER} columns show performance on the \textbf{AMI MixHeadset} \cite{carletta2005ami} and \textbf{DIHARD II} \cite{dihard} datasets, respectively.  
\textbf{Lower DER is better}, and (*) indicates results where overlapping speech regions were ignored.}}
\label{tab:combined_diarization_results}
\end{table*}

\subsection{Summarization} 
%Amina, Naome, Micheal
For open-source LLMs, we used publicly available checkpoints from HuggingFace \cite{wolf2020transformers} without altering their default hyperparameters, except for setting max\_new\_tokens to 1024. Closed-source models were accessed via their respective APIs, also using default hyperparameters. The prompt template was adapted from prior work \cite{zheng2023judging, liu2023benchmarking}, and to ensure consistency, the same prompt was used across all models (details can be found in the Appendix).

We conducted the summarization experiments under two scenarios: (1) generating summaries from human-produced transcripts and (2) generating summaries from transcripts created by our best-performing ASR model (Whisper-large-v3). %The experimental setup, including the consistent use of the prompt template, was maintained across both scenarios.

\section{Results and Discussion}

\subsection{Diarization}
\begin{figure}[!ht]
    \centering
    \begin{tikzpicture}[scale=0.5]
        \begin{axis}[
            ybar=10pt,
            width=\textwidth,
            height=0.5\textwidth,
            bar width=24pt,
            enlarge x limits=0.25,
            legend style={at={(0.5,-0.15)},
            anchor=north,legend columns=-1},
            ylabel={DER (\%)},
            symbolic x coords={Titanet-L, Pyannote, Reverb (reverb-diarization-v2)},
            xtick=data,
            nodes near coords,
            ymin=0,
            ymajorgrids=true,
            grid style=dashed,
            ticklabel style = {font=\large},
            ylabel style={font=\large},
            xlabel style={font=\large},
            title={Comparison of Medical and General DER for Different Models}
        ]
        \addplot+[draw=none,fill=cyan!70] 
            coordinates {(Titanet-L,34.64) (Pyannote,31.46) (Reverb (reverb-diarization-v2),58.04)};
        \addplot+[draw=none,fill=orange!70] 
            coordinates {(Titanet-L,12.28) (Pyannote,19.09) (Reverb (reverb-diarization-v2),20.10)};

        \legend{Medical DER (\%), General DER (\%)}
        \end{axis}
    \end{tikzpicture}
    \caption{Comparison of Medical and General DER for Different Models}
    \label{fig:derMed}
\end{figure}
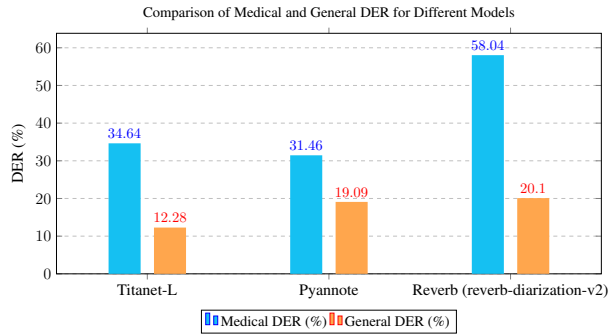

%The performance of the diarization models was evaluated using Diarization Error Rate (DER) across the timestamped 30 audio sample subset. We also computed the DER separately for medical (9 samples) and general (21 samples) domain conversations for a more granular analysis. The results are shown in Table \ref{tab:combined_diarization_results}. We also show the performance of these models on other conversational datasets, using the values reported by \cite{pyannote, titanet}.
We compute DER separately for a subset of conversations with accurate timestamps-- medical (9 samples, Med DER) and general (21 samples, Gen DER) domain conversations. The results are shown in Table \ref{tab:combined_diarization_results} and Figure \ref{fig:derMed}. We also show the performance of these models on conversational datasets on other accents, using the values reported in \cite{pyannote, titanet, vbx}.

The models consistently performed better on general domain conversations compared to medical conversations, likely due to their relaxed structure and fewer interruptions. Results show diarization results on Afrispeech-Dialog are better than AMI and DIHARD likely because of the simulated and structured nature of conversations.

%% Original Table 4 for DER%%%%%
% \begin{table*}[t]
% \centering
% \small
% %\resizebox{\textwidth}{!}{%
% \begin{tabular}{|l|c|c|c|c|c|}
% \hline
% \textbf{Model} & \textbf{DER (\%)} & \textbf{Med DER (\%)} & \textbf{Gen DER (\%)} & \textbf{AMI DER (\%)} & \textbf{DIHARD DER(\%)} \\ \hline
% Titanet-L & \textbf{16.27} & 34.64 & \textbf{12.28} & (1.89) & - \\ \hline
% Pyannote& 21.30 & \textbf{31.46} & 19.09 & 24.8 (4.6) & 34.4 \\ \hline
% Reverb & 26.87 & 58.04 & 20.10 & - & - \\ \hline
% \end{tabular}%
% %}
% \caption{Diarization Error Rate (DER) for all 30 audios, with detailed results for the Medical (Med. DER) and General (Gen. DER) subsets. The AMI DER and DIHARD DER columns show performance on the AMI MixHeadset \cite{carletta2005ami} and DIHARD II \cite{dihard} datasets, respectively. Lower DER is better, and (*) indicates results where overlapping speech regions were ignored.}
% \label{tab:combined_diarization_results}
% \end{table*}

% \begin{figure}
%     \centering
%     \includegraphics[width=1.2\linewidth]{fig/figure_a.pdf}
%     \caption{Comparison of Medical and General DER for Different Models}
%     \label{fig:derMed}
% \end{figure}

% \textcolor{red}{add graph, python/tikz (table-4)}

% also, medical conversation were significantly shorter, so error percentage is more pronounced.

\subsection{ASR}

We report WER for all conversations, as well as separately for the medical and non-medical portions of the data (Table \ref{tab:wer}). The \textbf{openai/whisper-large} models achieved the best performance, followed by \textbf{nvidia/canary}, \textbf{nvidia/parakeet} and  models and lastly \textbf{wav2vec2} based models. 

The results, as presented in Table \ref{tab:wer} and Appendix Figure \ref{fig:werMed}, demonstrate a clear trend: models exhibit superior performance on non-medical (general domain) audio compared to medical domain audio, roughly 5\% better WER, suggesting medical conversations are more challenging likely due to accented medical jargon such as medication names, diagnoses, and density of numerical entities \cite{afonja2024performant}.

Following trends in \citet{olatunji2023afrispeech}, Wav2vec2-large-960h finetuned exclusively on read librispeech audiobooks \cite{panayotov2015librispeech} performs worst, confirming that model size (number of parameters), training set size, training data domain (in-the-wild, conversational) and multilingual pretraining play a role in generalizability to accented speech.

Additionally, we compare performance on other conversational datasets reported in Table \ref{tab:wer} and show a 5 to 20 point (absolute) performance drop on average suggesting poor generalizability of SOTA models to African-accented conversation English speech. %consistently results in higher WERs than datasets featuring more widely studied accents. 
This degradation underscores the challenges ASR models face with African accents, highlighting the pressing need for high-quality, domain-specific datasets to improve ASR performance for low-resource accents and languages.

\begin{table*}[!ht]
\centering
\small
\renewcommand{\arraystretch}{1.1} % Adjust row height for better readability
\setlength{\tabcolsep}{6pt} % Adjust column spacing for better layout
\begin{tabular*}{\textwidth}{@{\extracolsep{\fill}} l c c c c c c}
\toprule  % Bold top horizontal line
\textbf{Model} & \textbf{WER} & \textbf{Med WER} & \textbf{Gen WER} & \textbf{AMI} & \textbf{Earnings22}  & \textbf{VoxPop} \\  
\midrule  % Single horizontal line for header separation
whisper-medium                      & 21.27   & 26.49           & \textbf{19.47}               & 16.6     & - & 7.4    \\  
whisper-large-v2                    & \textbf{20.82}   & \textbf{23.74}           & 19.81               & 16.4     & 12.05  & 7.3  \\  
whisper-large-v3                    & \textbf{20.38}   & \textbf{23.81}           & \textbf{19.19}               & 16.01 & \textbf{11.3} & - \\  
whisper-large-v3-turbo              & 21.93   & 25.58           & 20.67               & -     & -  & -  \\  
distil-whisper-large-v2             & 25.38   & 30.43           & 23.63               & \textbf{14.67}     & 12.19  & 8.24  \\  
distil-whisper-large-v3             & 21.20   & 25.67           & 19.58               & 15.95     & \textbf{11.29}  & 8.25  \\  
parakeet-rnnt-1.1b                  & 28.16   & 34.03           & 26.13               & 17.1  & 15.15 & \textbf{5.44} \\  
parakeet-ctct-1.1b                  & 28.97   & 34.16           & 27.19               & 15.67     & 13.75  & 6.56  \\  
parakeet-tdt-1.1b                   & 28.69   & 33.57           & 27.01               & 15.9  & 14.65 & \textbf{5.49} \\  
canary-1b                           & 22.82   & 27.40           & 21.25               & \textbf{13.53} & 12.05 & 5.79\\  
wav2vec2-large-960h               & 86.34   & 88.35           & 81.17               & 37.0     & -   & 17.9 \\  
mms-1b-all                        & 61.75   & 69.04           & 59.22               & 42.02     & 31.17  & 17.63  \\  
\bottomrule  % Bold bottom horizontal line
\end{tabular*}
\caption{\textit{WER for various pre-trained models on the entire dataset, including the \textbf{Medical} (Med WER) and \textbf{General} (Gen WER) portions, as well as AMI \cite{carletta2005ami}, Earnings22 \cite{del2022earnings}, and Voxpopuli \cite{wang-etal-2021-voxpopuli} datasets.  
\textbf{Lower is better}. Top 2 models in \textbf{bold}.}}
\label{tab:wer}
\end{table*}

\begin{table*}[!ht]
\centering
\small
\renewcommand{\arraystretch}{1.1} % Adjust row height for better readability
\setlength{\tabcolsep}{6pt} % Adjust column spacing for better layout
\begin{tabular*}{\textwidth}{@{\extracolsep{\fill}} l c c c c c}
\toprule  % Bold top horizontal line
\textbf{Model} & \textbf{BertScore-H} & \textbf{LLM-Eval-H} & \textbf{BertScore-M} & \textbf{LLM-Eval-M} & \textbf{Human-Eval} \\  
 & \textbf{(\%)} & \textbf{(\%)} & \textbf{(\%)} & \textbf{(\%)} & \textbf{(\%)} \\  
\midrule  % Single horizontal line for header separation
Claude-3-Sonnet & 88.39 & \textbf{73.53} & 88.50 & 69.42 & \textbf{69.74} \\  
GPT-4o & \textbf{91.34} & \textbf{72.94} & \textbf{90.94} & \textbf{70.98} & \textbf{67.48} \\  
GPT-3.5-turbo & 89.87 & 69.41 & 90.81 & 64.51 & - \\  
Llama-3.1-8B-Instruct & 86.44 & 69.41 & 86.44 & 62.35 & 65.87 \\  
Llama-3.2-3B-Instruct & 85.60 & 65.10 & 87.09 & 63.73 & 62.75 \\  
Gemma-2-9b & 86.00 & 71.37 & 85.48 & \textbf{71.57} & 60.83 \\  
Human Expert & - & - & - & - & 54.59 \\  
Llama3-Med42-8B & \textbf{90.56} & 67.45 & \textbf{91.34} & 65.29 & 51.11 \\  
Phi3-mini-instruct & 88.05 & 57.45 & 89.66 & 60.39 & 50.83 \\  
Meditron3-8B & 89.83 & 60.20 & 90.40 & 59.61 & 42.98 \\  
\bottomrule  % Bold bottom horizontal line
\end{tabular*}
\caption{\textit{Summarization performance of various models, evaluated using \textbf{BERTScore (F1)}, \textbf{LLM-Eval}, and \textbf{Human Evaluation}.  
\textbf{BERTScore-H} and \textbf{LLM-Eval-H} represent the metrics when summaries are generated and evaluated using \textbf{human-generated transcripts},  
while \textbf{BERTScore-M} and \textbf{LLM-Eval-M} represent the metrics for \textbf{machine-generated transcripts}.  
\textbf{Higher is better}. Top 2 models in \textbf{bold}.}}
\label{tab:summarization}
\end{table*}

% \begin{table*}[ht]
% \centering
% \small
% %\normalsize % Reduces font size of the table
% \begin{tabularx}{\textwidth}{|l|X|X|X|X|X|}
% \hline
% \textbf{Model} & \textbf{BertScore-H(\%)} & \textbf{LLM-Eval-H(\%)} & \textbf{BertScore-M(\%)} & \textbf{LLM-Eval-M(\%)} & \textbf{Human-Eval (\%) } \\
% \hline
% Claude-3-Sonnet & 88.39 & \textbf{73.53} & 88.50 & 69.42 & \textbf{69.74} \\
% \hline
% GTP4o & \textbf{91.34} & \textbf{72.94} & \textbf{90.94} & \textbf{70.98} & \textbf{67.48} \\
% \hline
% Gpt3.5-turbo & 89.87 & 69.41 & 90.81 & 64.51 & - \\
% \hline
% Llama-3.1-8B-Instruct & 86.44 & 69.41 & 86.44 & 62.35 & 65.87 \\
% \hline
% Llama-3.2-3B-Instruct & 85.60 & 65.10 & 87.09 & 63.73 & 62.75 \\
% \hline
% Gemma-2-9b & 86.00 & 71.37 & 85.48 & \textbf{71.57} & 60.83 \\
% \hline
% Human Expert & - & - & - & - & 54.59 \\
% \hline
% Llama3-Med42-8B & \textbf{90.56} & 67.45 & \textbf{91.34} & 65.29 & 51.11 \\
% \hline
% Phi3-mini-instruct & 88.05 & 57.45 & 89.66 & 60.39 & 50.83 \\
% \hline
% Meditron3-8B & 89.83 & 60.20 & 90.40 & 59.61 & 42.98 \\
% \hline
% \end{tabularx}
% \caption{Summarization performance of various models, evaluated using BERTScore (F1), LLM-Eval, and Human Evaluation. BERTScore-H and LLM-Eval-H represent the metrics when summaries are generated and evaluated using human-generated transcripts, while BERTScore-M and LLM-Eval-M represent the metrics for machine-generated transcripts. Higher is better. Top 2 models in Bold}
% \label{tab:summarization}
% \end{table*}

\begin{table*}[!ht]
\small
\centering
\begin{tabular}{lrrrrrrr}
\toprule
 Model & positives &  negatives &  diagnoses &  treatment &  no hallucination &  concise &  Total \\ \midrule
Human Expert & 3.35 & 3.06 & 1.97 &  1.59 & 3.23 & 3.18 &  16.37  \\
\hline
Claude-3-sonnet-20240229 & 4.31 & 4.31 & 1.54 & 1.46 & 4.77 & 4.54  &  \textbf{20.92} \\
GPT-4o & 4.22 & 4.13 & 1.71 & 1.76 & 4.31 & 4.11  &  20.24 \\
Llama-3.1-8B-Instruct & 3.90 & 3.71 & 2.67 & 2.14 & 3.71 & 3.62  &  19.76 \\
Llama-3.2-3B-Instruct & 4.04 & 3.91 & 1.35 & 1.43 & 3.96 & 4.13  &  18.83 \\
Gemma & 3.81 & 3.88 &  1.31 &  1.69 & 3.69 &  3.88 &  18.25 \\
Llama3-Med42-8B & 3.13 &  2.67 & 2.00 & 1.87 &   2.83 & 2.83 &  15.33  \\
Phi-3-mini-4k-instruct & 3.22 & 2.84 & 1.59 & 1.67 &  2.96 &  2.96  &  15.25 \\
Meditron3-8B & 2.86 & 2.48 & 1.46 & 1.23 & 2.45 &  2.41  &  12.89 \\
\bottomrule
\end{tabular}
\caption{Human Evaluation Results showing means of 5-point ratings for each criteria. Higher is better. \textbf{Positives:} \textit{Does the summary include all important positive clinical signs, symptoms, or details in the transcript?};  \textbf{Negatives:} \textit{Does the summary include all important negative clinical signs, symptoms, or details in the transcript?}; \textbf{Diagnoses:} \textit{Does the summary accurately recall the diagnoses from the transcript?}; \textbf{Treatment:} \textit{Does the summary accurately reflect the treatment plan in the transcript?}; \textbf{No Hallucination:} \textit{Does the summary avoid any false- incorrect- or fabricated information?}; \textbf{Concise:} \textit{Is the summary concise- clear- and well-structured?}
}
\label{tab:human-eval}
\end{table*}

\subsection{Summarization}
%Amina, Naome, Micheal
%In this section, we present the results of our summarization experiments, focusing on the comparison between human and model-generated summaries across various scenarios. 
Table \ref{tab:summarization} shows our summarization results.

In BERTScore-H, summaries are generated based on human transcripts and evaluated against human reference summaries. In BERTScore-M, summaries are generated based on machine-predicted transcripts (Whisper-Large-v3) and evaluated against human reference summaries. 

LLM-Eval-H represents the LLM-as-Judge results when key details in the human transcripts are sought from summaries generated from human transcripts. LLM-Eval-M represents the LLM-as-Judge results when key details in the human transcripts are sought from summaries generated from machine-predicted transcripts.

Table \ref{tab:summarization} shows that general domain models like GTP4o and Claude consistently achieve high scores across all evaluation metrics followed by Biomedical LLMs where generated summaries are semantically more similar to reference summaries.
Claude-3 was ranked highest for summarizing human transcripts, while Gemma-2 ranked highest for summarizing ASR transcripts.  
%BERTScore-H and LLM-Eval-H represent the metrics when summaries are generated and evaluated using human-generated transcripts, while BERTScore-M and LLM-Eval-M represent the metrics for machine-generated transcripts.

%In contrast, the medical models performed slightly lower, indicating the need for further domain-specific fine-tuning to handle diverse summarization tasks better. 

% The Llama models delivered mixed results, with Meta-Llama-3.1-8B-Instruct and Meta-Llama-3.2-3B-Instruct maintaining consistent BERTScore results, while their performance on LLM-Eval metrics, especially for ASR-generated transcripts, was lower. The weakest performer overall was Phi3-mini-4k-instruct, particularly in the LLM-Eval scores, signaling significant challenges in medical summarization.

Open general smaller LLMs like Gemma and the Llama models demonstrate lower semantic similarity with reference summaries (BERTScore) and their performance on LLM-Eval metrics demonstrate an inferior ability to retrieve critical information from the conversation. The weakest performer overall was Phi3-mini-4k-instruct, particularly in the LLM-Eval scores, signaling significant challenges in medical summarization.

%Table \ref{tab:summ_info_loss} presents a comparison between a human-written transcript and the corresponding summary generated by GPT4o. While GPT4o captures key details such as pain severity, lifestyle factors, and medical history, its summary omits a vital detail: the patient’s experience of rectal bleeding ``during urination", which is crucial to the case. For a complete presentation of the case and summaries generated by other models, please refer to the Appendix section \ref{sec:summ_examples}.

% OpenMeditron-Meditron3-8B

\subsection{Human Evaluation}
Results in Table \ref{tab:summarization} show that LLM-as-Judge ratings (LLM-Eval-H) were strongly correlated (Pearsons=0.816) with Human Expert ratings, in line with \citet{zheng2023judging}. 

Table \ref{tab:human-eval} showed that, of the 6 criteria, expert summaries were ranked in the top 3 only at accurately recalling the diagnoses from the transcript, suggesting this was a key focus for experts. LLM summaries were more concise and retained greater amounts of key details when compared to the expert summaries. 

The blind expert ratings showed that overall, experts ranked leading LLM summaries higher than reference summaries, scoring consistently higher on the completeness of key facts carried over from the transcript to the summary. This suggests that LLMs may perform better at certain tasks, such as compressing large amounts of information (10+ mins conversations) when compared with time-constrained physicians who may prefer brevity. 

\subsection{Error Propagation from cascading models}

%When comparing the models on BERTScore-H and BERTScore-M, no significant differences emerged, suggesting that most models handle both human and ASR transcripts similarly at the token level. However, a more nuanced pattern appears with LLM-Eval-H and LLM-Eval-M metrics. Here, many models show a notable drop in performance when evaluated on ASR transcripts, with lower LLM-Eval-M scores. This indicates that noise and transcription errors in ASR outputs pose challenges for models, particularly when assessed with comprehensive evaluation methods that measure summary quality and coherence beyond text overlap.

While summarizing ASR transcripts reveal no significant difference in semantic similarity with reference summaries (BERTScore-H vs BERTScore-M), a more nuanced pattern emerges with LLM-Eval-H and LLM-Eval-M metrics where many models show a notable drop in performance. LLM-Eval-M scores are on average 2 to 5 points lower than LLM-Eval-H. This indicates that ASR errors (noisy transcription outputs) may limit the summarization ability of LLMs leading to loss of critical healthcare information. 

%The Gemma demonstrates a greater resilience to ASR noise, exhibiting only minor or no performance decline between LLM-Eval-H and LLM-Eval-M, suggesting its suitability for noisy input tasks. In contrast, models like OpenMeditron-Meditron3-8B and Phi3-mini-4k-instruct struggle in these conditions, particularly on LLM-Eval metrics, highlighting their limitations in handling noisy inputs.

% Diet: The mention of "swallow" in the diet is accurate, but more context could be provided (e.g., it being a staple in the patient's diet).

% making it the best performer. GPT-3.5-turbo and Claude_Sonnet are also notable contenders, though they lag in LLM-Eval scores for ASR.

% We compared the summaries from all nine models against gold-standard summaries written by medical experts. These expert-generated summaries ensured that the evaluations were grounded in domain-specific knowledge, making the BERT Score results highly relevant in assessing how well the models preserved medical meaning. This evaluation offered insights into the models' ability to condense dialogues while maintaining essential medical information, particularly in a multilingual context. 

% In this section, we present the results of our summarization experiments, focusing on the comparison between human and model-generated summaries across different scenarios. 

\section{Conclusion}
This study highlights several challenges with accented conversational ASR, especially its impact on downstream medical summarization. We benchmark speaker diarization, ASR, and medical summarization, drawing attention to gaps and opportunities to improve accented conversational ASR in the African context.

\section{Limitations}
% Lukman, Tobi
% Limitation: simulated vs real, age range is contributors is not reflective of real-world age distr of speakers; use of general LLM as judge
% Ethics, releasing real voices, voice cloning biometric, but identities were concealed with pseudonyms, small dataset size

% n addition, although the OSCE- styled medical conversations are superior to traditional clinical exams in terms of objectivity, precision, and ability to handle unpredictable patient behavior and seemingly unrelated symptoms, they are limited in their ability to simulate real-world patient-physician conversations, which are more complex due to subtle body language, facial cues and other non-verbal presentations. Thirdly, these medical conversations only covered the history-taking part of simulated medical visits. Physical exams were not included in the medical conversation and therefore, there may be limitations in informing a clinical differential diagnosis and management plan. This dataset has 3309 minutes of audio and 272 transcribed texts. Training AI models is data-intensive requiring large amounts of data25,26,27. Therefore, this dataset can be combined with other datasets for the purposes described previously. The user will have to take into consideration transferability and generalizability when combining such data. Lastly, this dataset focussed predominantly on respiratory cases so it does limit usage. 

While simulated conversations provide the opportunity to benchmark ASR systems, they may not fully reflect the complexity of real-world natural dialogues where several interruptions, distractions and background noise are more typical. Additionally, the age range of contributors does not accurately reflect the diversity of real-world speaker populations, potentially impacting the generalizability of the results. The use of general large language models (LLMs) as evaluators may also introduce biases. At 5hrs, the dataset does not reflect the full breadth of medical or general conversations as several domains, subdomains, topics, and medical specialties were not covered in this release.  %Moreover, computational and data-access constraints limited the exploration of more diverse linguistic settings and model fine-tuning. 
Nonetheless, this work significantly contributes to addressing the gap in ASR systems' adaptability to African accents. Future research should aim to incorporate real-world data, broader age representation, and further optimize models to enhance the robustness and applicability of ASR technologies in diverse African linguistic environments.

\section{Ethical Considerations}
%Ethics, releasing real voices, voice cloning biometric, but identities were concealed with pseudonyms, small dataset size
In a world where voice cloning and identity theft are major cybersecurity threats, releasing a voice-based dataset is not without risk. Each recording began with contributors expressing consent to the recording. However, since real names were sometimes used during consent, these segments were removed to protect participants' identities. During doctor-patient conversations, pseudonyms were used to maintain anonymity. Furthermore, releasing real-world doctor-patient conversations risks breaching privacy and confidentiality requirements, hence the preference for simulated conversations. 

%Ethical considerations were central to this work, as the dataset relies on simulated conversations to maintain anonymity, with contributors’ identities protected using pseudonyms, thereby mitigating potential biometric risks associated with voice cloning. 

\section{Acknowledgment}
We thank the Intron Health team for providing the data and compute used in this work and all contributors, patient- and doctor-actors whose voices are included in the dataset.

%\subsection{References}

\nocite{Ando2005,andrew2007scalable,rasooli-tetrault-2015}

%\section*{Acknowledgments}

% Bibliography entries for the entire Anthology, followed by custom entries
%\bibliography{anthology,custom}
% Custom bibliography entries only
\bibliography{latex/main}

\newpage
\appendix

\section{Human Evaluation Criteria} 
\label{sec:human_Eval}
% \vspace{-30pt} 
This section provides details on the criteria used to evaluate the AI-generated summaries of doctor-patient conversations. Each summary was evaluated based on the following aspects: 

\begin{enumerate}
 \setlength{\itemsep}{0pt}
    \setlength{\parskip}{0pt}
    \setlength{\parsep}{0pt}
    \item  How well does the summary include all positive clinical signs, symptoms, or details in the transcript?
    \item How well does the summary include all negative clinical signs, symptoms, or details in the transcript?
    \item How well does the summary recall the diagnosis from the transcript?
    \item How accurately does the summary reflect the treatment plan in the transcript?
    \item  To what degree does the summary avoid false, incorrect, or fabricated information?
    \item  How concise, clear, and well-structured is the summary?
\end{enumerate}

\section{Prompt Template for the LLM Evaluator}
\label{sec:llm_eval_prompt}

Table \ref{tab:llm_evaluation_criteria} outlines the prompt template utilized for the evaluation conducted by OpenAI's o1 model, serving as the LLM evaluator.

\begin{table*}[htbp]
\centering
\renewcommand{\arraystretch}{1.2} % Adjust row height
\begin{tabular}{|p{0.9\textwidth}|}
\hline
\textbf{LLM as a Judge Prompt Template} \\ \hline
You are a medical expert. Your task is to evaluate the accuracy and quality of an AI-generated summary of a doctor-patient conversation. Provide an impartial assessment based on the criteria below, rating each from 1 to 5 (1 = lowest, 5 = highest). If a specific criterion is not addressed, assign a score of 0. \\ 

\textbf{Transcript:} \{transcript\} \\ 
\textbf{Summary:} \{summary\} \\

\textbf{Evaluation Criteria:} \\ 
1. Completeness of positive clinical signs, symptoms, or relevant details: Does the summary capture all important positive findings? \\
2. Completeness of negative clinical signs, symptoms, or relevant details: Does the summary capture all important negative findings? \\
3. Accuracy of diagnosis: Does the summary correctly reflect the diagnosis? \\
4. Accuracy of treatment plan: Does the summary convey the treatment plan correctly? \\
5. Factual accuracy: Does the summary avoid false, incorrect, or fabricated information? \\
6. Clarity and structure: Is the summary concise, clear, and well-structured? \\ 
\textbf{Please Return the evaluation scores in JSON format:} \\
\{ \\
\hspace{1em} "Criterion 1": score, \\
\hspace{1em} "Criterion 2": score, \\
\hspace{1em} "Criterion 3": score, \\
\hspace{1em} "Criterion 4": score, \\
\hspace{1em} "Criterion 5": score, \\
\hspace{1em} "Criterion 6": score \\
\} \\ \hline
\end{tabular}
\caption{Evaluation task description for assessing AI-generated summaries.}
\label{tab:llm_evaluation_criteria}
\end{table*}

\section{Evaluation Metric for the LLM Evaluator}
\label{sec:llm_metric}

In the LLM evaluation prompt (see Section~\ref{sec:llm_eval_prompt}), each criterion has a maximum score of 5 points. Since there are six criteria per row, the perfect score for a single row is \(5 \times 6 = 30\) points. For a dataset with ten rows, the total possible score (perfect accuracy) is:
\vspace{-10pt} 

\[
\text{Perfect Accuracy} = 5 \times 6 \times 10 = 300 \, \text{points}.
\]

To compute a model's percentage score for a dataset with N rows:
\begin{enumerate}
    \setlength{\itemsep}{0pt}
    \setlength{\parskip}{0pt}
    \setlength{\parsep}{0pt}
    \item Sum the scores for all six criteria per row.
    \item Sum the total scores across all rows.
    \item Calculate the percentage score for the model as:
\end{enumerate}

\small
\[
\text{Percentage Score} = 
\left( \frac{\text{Model's Total Score}}{5 \times 6 \times \text{Number of Rows}} \right) \times 100.
\]
\normalsize

\vspace{1em} % Adds space between sections

\section{Prompt Template for Summarization}
\label{sec:summ_prompt}
This figure outlines the prompt template utilized for the summarization task.
\\

\begin{minipage}{\linewidth} % Create a minipage to allow text box across columns

\noindent\hspace{-0.5cm}%
\fbox{%
    \parbox{\textwidth}{%
        \small
        You are a medical expert; you will be provided a transcript that contains a conversation between a doctor and a patient. The task is for you to summarize the key points in the conversation.
\textbf{Transcript:} \{transcript\}
    }
}
\end{minipage}

\section{Transcript and Summary Example}
\label{sec:summ_examples}
This section presents a sample of a human-written transcript in Table \ref{tab:summ_info_loss} and the corresponding LLM-generated summaries for the transcript. While GPT4o captures key details such as pain severity, lifestyle factors, and medical history, its summary omits a vital detail: the patient’s experience of rectal bleeding ``during urination", which is crucial to the case. The summaries are:\\

\begin{table*}[h!]
\centering
\begin{tabularx}{\textwidth}{|X|}
\hline
\textbf{Transcript} \\ 
\hline
\textbf{Doctor:} Good afternoon. My name is Dr. Philip, and what's your name? \\ 
\hline
\textbf{Patient:} My name is Mrs. Ayike. \\ 
\hline
\textbf{Doctor:} Okay, how old are you? \\ 
\hline
\textbf{Patient:} I am 45 years old. \\ 
\hline
\textbf{Doctor:} Okay. You're welcome, my dear. What brought you here? \\ 
\hline
\textbf{Patient:} I have rectal bleeding and abdominal pain. \\ 
\hline
\textbf{Doctor:} Can you tell me more about it? \\ 
\hline
\textbf{Patient:} The pain started two weeks ago, and the rectal bleeding happens when I urinate. The blood is not mixed with stool. \\ 
\hline
\textbf{Doctor:} Have you gone for any check-ups? \\ 
\hline
\textbf{Patient:} No, this is my first time coming here. \\ 
\hline
\textbf{Doctor:} How severe is the abdominal pain, from a scale of 1 to 10? \\ 
\hline
\textbf{Patient:} 6. \\ 
\hline
\textbf{Doctor:} Which side of the abdominal region do you feel the pain? \\ 
\hline
\textbf{Patient:} I feel pain in my lower abdominal region, and it worsens with pressure. And since the last two weeks, as I've started the abdominal pain, I have not been able to eat well. I lost appetite. I feel tired. \\ 
\hline
\textbf{Doctor:} Have you traveled recently? \\ 
\hline
\textbf{Patient:} No, I don't travel. \\ 
\hline
\textbf{Doctor:} Are you married? \\ 
\hline
\textbf{Patient:} Yes, I am married, with four children, but my husband lives in another state. \\ 
\hline
\textbf{Doctor:} Have you undergone any surgery before? \\ 
\hline
\textbf{Patient:} No. \\ 
\hline
\textbf{Doctor:} Do you smoke or drink alcohol? \\ 
\hline
\textbf{Patient:} No. \\ 
\hline
\textbf{Doctor:} Does anyone in your family have similar issues? \\ 
\hline
\textbf{Patient:} Yes, my mother had colorectal cancer. \\ 
\hline
\textbf{Doctor:} What foods do you normally eat? \\ 
\hline
\textbf{Patient:}  I like taking swallow. I like swallow a lot. \\ 
\hline
\textbf{Doctor:} Are you allergic to any food? \\ 
\hline
\textbf{Patient:} No. \\ 
\hline
\end{tabularx}
\caption{Snippet of a doctor-patient conversation (transcript) written by a clinician}
\label{tab:summ_info_loss}
\end{table*}

\textbf{GPT4o's Summary:} Dr. Philip conducted a consultation with Mrs. Ayike, a 45-year-old woman. She reported experiencing rectal bleeding and abdominal pain for the past two weeks. The pain, rated as a 6 out of 10 on the pain scale, is localized in the lower abdominal region and worsens with pressure. Mrs. Ayike also noted a loss of appetite and fatigue since the onset of her symptoms. She has not sought medical attention prior to this visit and has no history of surgery. She neither smokes nor drinks alcohol. Although Mrs. Ayike does not travel, she revealed a family history of colorectal cancer through her mother. Her usual diet includes "swallow," and she has no food allergies. Mrs. Ayike is married with four children, but currently lives separately from her husband. Dr. Philip concluded the session by acknowledging Mrs. Ayike's details. \\

\textbf{Claude's Summary:} Based on the conversation, the key points are: 1. The patient, Mrs. Ayike, is a 45-year-old woman presenting with rectal bleeding and abdominal pain for the past two weeks. 2. The rectal bleeding occurs when she urinates, and it is not mixed with stool.3. The abdominal pain is rated as 6 out of 10 in severity, and it is localized in the lower abdominal region.4. She has lost her appetite and feels tired since the symptoms started.5. She has not traveled recently and is not a frequent traveler.6. She is married but currently separated from her husband, living in a different state.n7. She has four children.8. She has not undergone any surgery before. 9. She does not smoke or consume alcohol.10. Her mother had colorectal cancer in the past, indicating a family history of the condition.11. Her typical diet includes swallows (a type of food), and she is not allergic to any food. The doctor has gathered essential information about the patient's symptoms, medical history, and lifestyle to aid in the diagnosis and treatment process. \\

\textbf{Gemma's Summary:} Mrs. Ayike, a 45-year-old woman, presented to Dr. Philip complaining of two weeks of abdominal pain and rectal bleeding. 
Key points:Symptoms:Lower abdominal pain rated 6/10, worsening with pressure.
Rectal bleeding during urination, not mixed with stool. Loss of appetite and fatigue. Medical History: No previous surgeries. No smoking or alcohol consumption. Family history of colorectal cancer (mother). Lifestyle: Prefers a diet high in swallow. No known food allergies. Not a traveler. Separated from her husband, living with four children. Next steps: Based on the symptoms and family history, Dr. Philip will likely order further investigations to determine the cause of Mrs. Ayike's abdominal pain and rectal bleeding. This may include blood tests, stool tests, imaging studies (such as a colonoscopy or CT scan), and possibly a biopsy.
\\

\textbf{Med42-8B's summary:} Mrs. Ayike, a 45-year-old female, presents with rectal bleeding and lower abdominal pain, rated 6/10 in severity, which started two weeks ago. She reports seeing red blood in her urine but not mixed with stool. She has not sought medical attention previously for these symptoms. She describes her appetite as poor and feels tired. She denies any recent travel, smoking, or alcohol use. She is married with four children but is currently separated from her husband. Her mother had colorectal cancer. She consumes a diet primarily consisting of fufu (swallow). There are no known food allergies.
\\

\textbf{GPT3.5's summary:} Mrs. Aike, a 45-year-old patient, visited Dr. Philip due to rectal bleeding and abdominal pain that started two weeks ago. The abdominal pain is severe, rated at a 6 on a scale of 1 to 3, and is localized in the lower abdominal region. She has also experienced a loss of appetite, fatigue, and difficulty eating since the pain started. Mrs. Aike is married with four children, not living with her husband, and has no history of surgeries, smoking, or alcohol consumption. Her mother had colorectal cancer before, indicating a family history of similar issues. She prefers eating swallow and is not allergic to any food. This is her first visit to seek medical help for these symptoms.
\\

\textbf{Meditron3-8B's summary:} The patient, Mrs. Ayike, is a 45-year-old woman who has been experiencing rectal bleeding and abdominal pain for two weeks. She has not undergone any surgery before and does not smoke or take alcohol. Her mother had colorectal cancer before. She has lost her appetite and feels tired. The doctor asks her about her family history, food allergies, and the severity of her symptoms. The doctor also asks about her marital status and children. The patient's symptoms are similar to those of colorectal cancer, which her mother had.
\\

\textbf{Llama-3.1-8B-Instruct's summary:} Based on the provided transcript, here are the key points from the conversation between Dr. Philip and Mrs. Ayike: 1. Patient's Introduction: Mrs. Ayike, a 45-year-old woman, presents to the doctor with rectal bleeding and abdominal pain. 2.Symptoms: The abdominal pain started two weeks ago, and she experiences rectal bleeding when urinating, but not mixed with stool. The pain is located in the lower abdominal region and worsens with pressure. 3. Medical History: Mrs. Ayike has not undergone any surgery before and does not have a history of smoking or alcohol consumption. 4. Family History: Her mother had colorectal cancer before.5. Diet and Allergies: Mrs. Ayike likes eating swallow (a type of food) and is not allergic to any food. 6.Lifestyle : She is married with four children but is currently separated from her husband. She does not travel and has not been able to eat well or feel energetic due to her symptoms. 7. Current Condition: Mrs. Ayike rates her abdominal pain as a 6 out of 10 and has lost appetite and feels tired since the onset of symptoms.
Based on the information provided, Dr. Philip will likely perform a thorough examination and possibly order diagnostic tests to determine the cause of Mrs. Ayike's symptoms. Given her family history of colorectal cancer, a colonoscopy may be recommended to rule out any potential malignancies.
\\

\textbf{Llama-3.2-3B-Instruct's summary:} Here's a summary of the key points in the conversation between Dr. Philip and Mrs. Ayike: Patient's Symptoms: Rectal bleeding (not mixed with stool), Abdominal pain (rated 6 out of 10) in the lower abdominal region, Loss of appetite and tiredness for the past two weeks. Medical History: No previous surgeries, No smoking or alcohol consumption, No known allergies to food Family History: Mother had colorectal cancer before. Lifestyle and Habits: Married with four children, but not living with her husband. No history of traveling. Prefers swallowing food. Other Information: Mrs. Ayike is 45 years old. She has not been able to eat well due to the abdominal pain. She is seeking medical attention for the first time, as the symptoms started two weeks ago.
\\

\textbf{Phi3-mini-instruct's summary:}  Dr. Philip, a medical expert, is speaking with Mrs. Ayike, a 45-year-old woman who has been experiencing rectal bleeding and abdominal pain for two weeks. Mrs. Ayike has not had any check-ups for her symptoms and has lost her appetite, feeling tired. She has not traveled recently and is not a smoker. She does not consume alcohol and has no known food allergies. Mrs. Ayike's mother had colorectal cancer. Dr. Philip asks about her diet, which mainly consists of swallow, and inquires about any family history of similar issues.

\section{Overlapping Speech Example}
\begin{table*}[h!]
\centering
\begin{tabular}{|c|l|p{10cm}|}
\hline
\textbf{Timestamp} & \textbf{Speaker} & \textbf{Utterance} \\ \hline
02:40.98 -- 02:41.99 & Speaker 1 & Past hospital admission? \\ \hline
02:42.97 -- 02:43.02 & Speaker 2 & No. \\ \hline
02:43.00 -- 02:45.99 & Speaker 1 & Ok, thank you very much. Do you live alone? \\ \hline
\end{tabular}
\caption{Example of overlapping speech with accurate timestamps and speaker annotations.}
\label{tab:overlap_example}
\end{table*}

In this example, overlapping speech occurs between Speaker 1 and Speaker 2. Speaker 2’s "No" overlaps with Speaker 1’s utterance starting with "Ok, thank you very much." Both utterances are fully captured with precise timestamps to ensure accurate representation of the conversation. This example illustrates how overlaps are handled in the transcription process for the dataset.

\section{ASR and Summarization Result Graphs}

%\textcolor{red}{add graph group bar-chart, (table-6) python/tikz package}

\begin{figure*}
    \centering
    \includegraphics[width=1\linewidth]{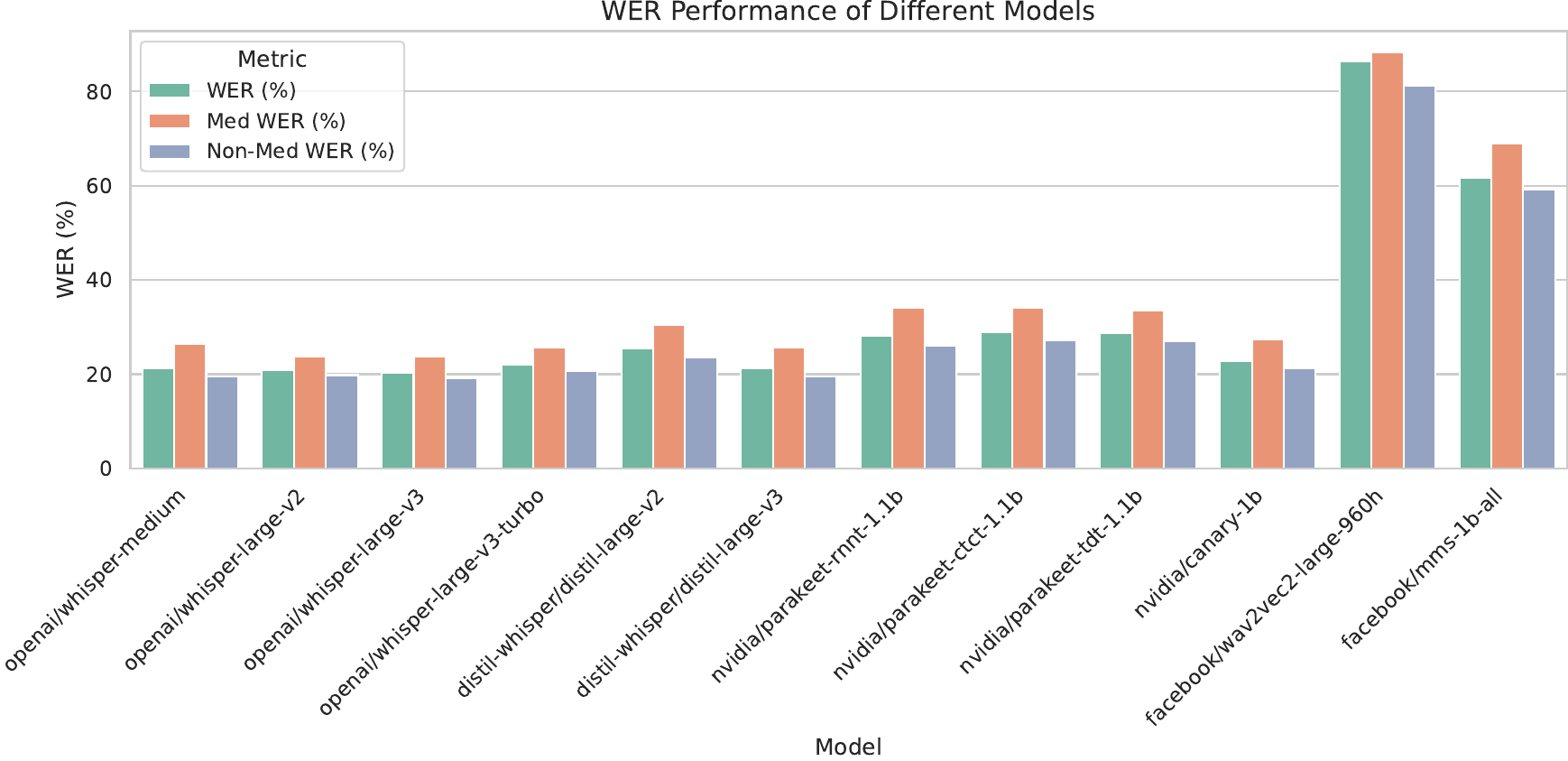}
    \caption{Comparison of Medical and General WER for Different Models}
    \label{fig:werMed}
\end{figure*}

\begin{figure*}
    \centering
    \includegraphics[width=1\linewidth]{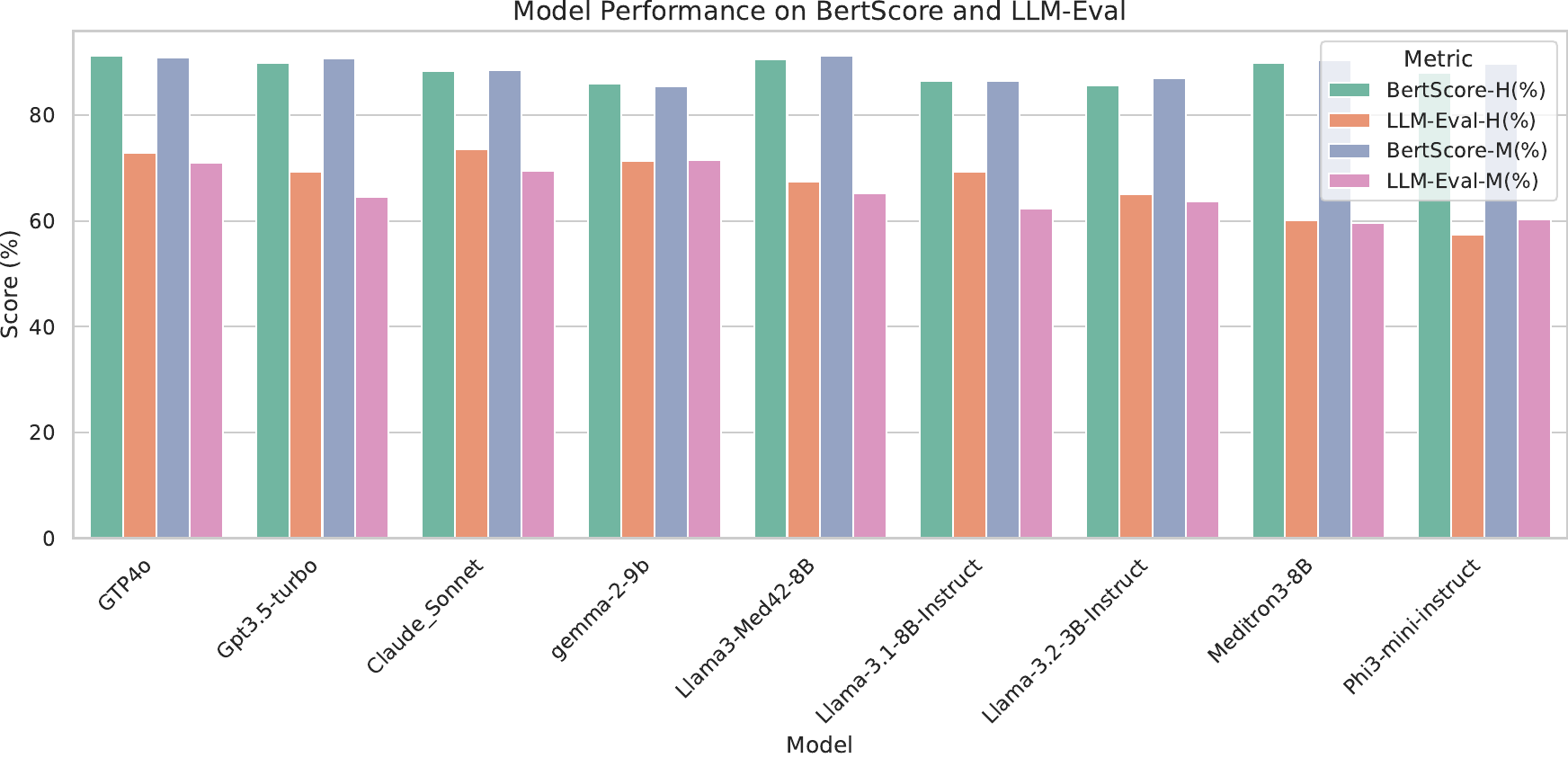}
    \caption{Summarization results for several LLMs}
    \label{fig:summ_Med}
\end{figure*}

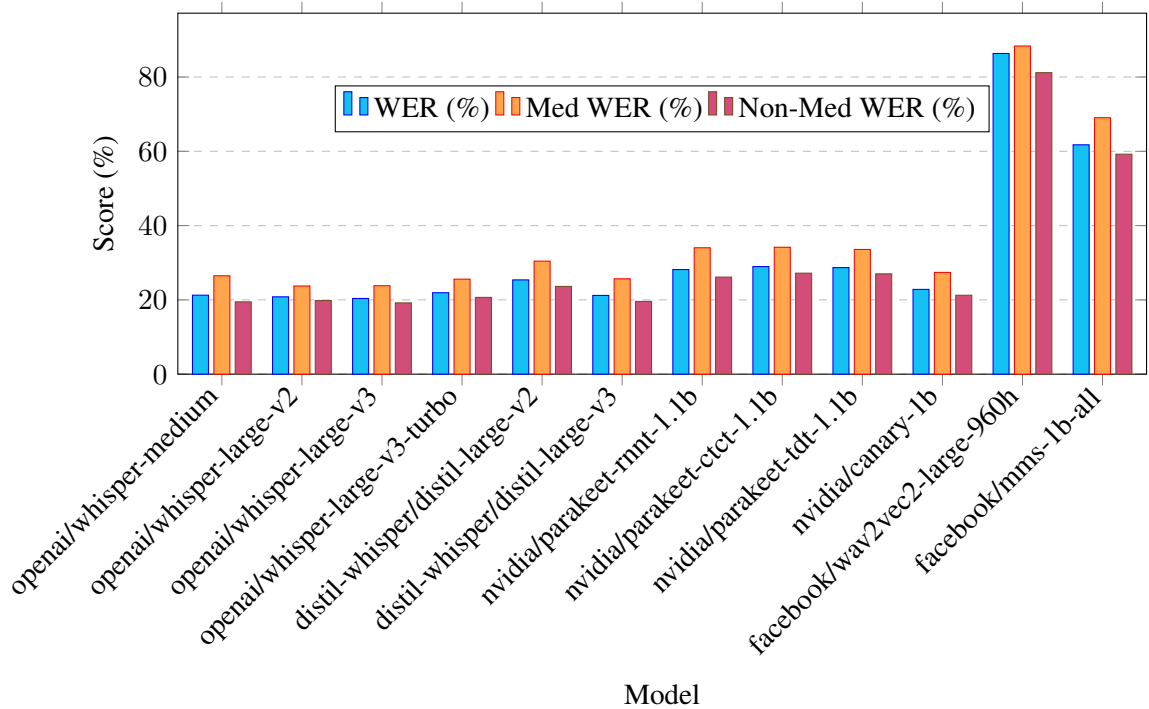
\begin{figure*}[ht]
    \centering
    \begin{tikzpicture}
        \begin{axis}[
            ybar,
            bar width=6pt,
            width=0.9\textwidth,
            height=0.4\textwidth,
            xlabel={Model},
            ylabel={Score (\%)},
            symbolic x coords={openai/whisper-medium, openai/whisper-large-v2, openai/whisper-large-v3, 
                               openai/whisper-large-v3-turbo, distil-whisper/distil-large-v2, 
                               distil-whisper/distil-large-v3, nvidia/parakeet-rnnt-1.1b, 
                               nvidia/parakeet-ctct-1.1b, nvidia/parakeet-tdt-1.1b, 
                               nvidia/canary-1b, facebook/wav2vec2-large-960h, facebook/mms-1b-all},
            xtick=data,
            enlarge x limits=0.05,
            xticklabel style={rotate=45, anchor=east},
            ymin=0,
            legend style={at={(0.5,0.8)}, anchor=north,legend columns=-1},
            ymajorgrids=true,
            grid style=dashed,
        ]
        
        % WER (%) bars
        \addplot+[ybar, fill=cyan!70] 
            coordinates {(openai/whisper-medium, 21.27) (openai/whisper-large-v2, 20.82) 
                         (openai/whisper-large-v3, 20.38) (openai/whisper-large-v3-turbo, 21.93) 
                         (distil-whisper/distil-large-v2, 25.38) (distil-whisper/distil-large-v3, 21.20) 
                         (nvidia/parakeet-rnnt-1.1b, 28.16) (nvidia/parakeet-ctct-1.1b, 28.97) 
                         (nvidia/parakeet-tdt-1.1b, 28.69) (nvidia/canary-1b, 22.82) 
                         (facebook/wav2vec2-large-960h, 86.34) (facebook/mms-1b-all, 61.75)};
        \addlegendentry{WER (\%)}
        
        % Med WER (%) bars
        \addplot+[ybar, fill=orange!70] 
            coordinates {(openai/whisper-medium, 26.49) (openai/whisper-large-v2, 23.74) 
                         (openai/whisper-large-v3, 23.81) (openai/whisper-large-v3-turbo, 25.58) 
                         (distil-whisper/distil-large-v2, 30.43) (distil-whisper/distil-large-v3, 25.67) 
                         (nvidia/parakeet-rnnt-1.1b, 34.03) (nvidia/parakeet-ctct-1.1b, 34.16) 
                         (nvidia/parakeet-tdt-1.1b, 33.57) (nvidia/canary-1b, 27.40) 
                         (facebook/wav2vec2-large-960h, 88.35) (facebook/mms-1b-all, 69.04)};
        \addlegendentry{Med WER (\%)}
        
        % Non-Med WER (%) bars
        \addplot+[ybar, fill=purple!70] 
            coordinates {(openai/whisper-medium, 19.47) (openai/whisper-large-v2, 19.81) 
                         (openai/whisper-large-v3, 19.19) (openai/whisper-large-v3-turbo, 20.67) 
                         (distil-whisper/distil-large-v2, 23.63) (distil-whisper/distil-large-v3, 19.58) 
                         (nvidia/parakeet-rnnt-1.1b, 26.13) (nvidia/parakeet-ctct-1.1b, 27.19) 
                         (nvidia/parakeet-tdt-1.1b, 27.01) (nvidia/canary-1b, 21.25) 
                         (facebook/wav2vec2-large-960h, 81.17) (facebook/mms-1b-all, 59.22)};
        \addlegendentry{Non-Med WER (\%)}

        \end{axis}
    \end{tikzpicture}
    \caption{WER, Med WER, and Non-Med WER for Various Models}
\end{figure*}

\end{document}